\documentclass[letterpaper,10pt,conference]{ieeeconf}
\usepackage{graphicx}
\usepackage{amsmath}
\usepackage{url}

\title{\LARGE \bf
Toward Evidence-Driven Human-Agent-Robot Teaming\\
for Earth-Independent Anomaly Triage}

\author{
Ignacio G. Lopez-Francos$^{1,2,3}$,
Samira Shalal$^{1,4}$,
and Alexis Gallagher$^{5}$
}

\begin{document}
\maketitle

\begingroup
\renewcommand{\thefootnote}{\arabic{footnote}}

\footnotetext[1]{NASA Ames Research Center, Moffett Field, CA, USA}
\footnotetext[2]{SETI Institute, Mountain View, CA, USA}
\footnotetext[3]{The University of Texas at Austin, Austin, TX, USA}
\footnotetext[4]{University of Michigan, Ann Arbor, MI, USA}
\footnotetext[5]{Repertory Robotics, San Francisco, CA, USA}
\endgroup

\thispagestyle{empty}
\pagestyle{empty}

\begin{abstract}
Deep-space crews cannot rely on real-time ground support for urgent off-nominal events. Initial alerts may underdetermine cause, while discriminating evidence may reside in crew observations or at locations that are unsafe, costly, or unavailable for crew inspection. We present an evidence-driven architecture for human-agent-robot teaming in Earth-independent anomaly triage. Agentic AI is treated as a coordinator over explicit state and bounded, inspectable services rather than as a fully autonomous vehicle controller. A triage state manager maintains hypotheses, evidence provenance, uncertainty, operational context, and tool status; a crew-facing embodied agent elicits observations and explains assessment changes; and a mobile robot acquires targeted, localized evidence. Typed interfaces separate dialogue and orchestration from monitoring, robot command, context retrieval, and safety-critical control. Two scenarios illustrate the architecture: a crewed deep-space mission based on an actual ISS ammonia false alarm, where suspected contamination restricts crew access, and a power-interface anomaly at a crewed lunar base, where robotic inspection distinguishes a local connector fault from other causes ambiguous in remote telemetry. Our contribution is an authority-bounded closed evidence-loop architecture, exercised in a hardware-in-the-loop integration prototype using Reachy Mini and an Innate MARS mobile robot.
\end{abstract}

\section{Introduction}

As human missions extend beyond low-Earth orbit, crews will increasingly conduct time-critical operations without real-time Mission Control support. Mars missions can experience one-way communication delays of up to 22 minutes and extended disruptions, precluding synchronous troubleshooting \cite{chappell2024,mcbrayer2022}. Earth-independent operations therefore require onboard support for time-critical anomaly response, not merely onboard copies of ground displays \cite{maddox2024}.

An initial alert may indicate a serious condition without resolving its cause or safest response. The next step may therefore be a targeted observation from vehicle telemetry, the crew, a deployable sensor, or a mobile robot rather than an immediate repair attempt. Treating the robot as a relocatable sensing resource follows active perception, in which sensing actions are selected to reduce uncertainty \cite{bajcsy1988}, and extends targeted off-world fault investigation \cite{gorospe2026} by closing the loop from an unresolved distinction to robot tasking, structured evidence return, hypothesis revision, and crew review.

Tool-using language agents can interleave reasoning and external action \cite{yao2023} and ground instructions in finite robot skill sets \cite{ichter2023}, but safety-critical operations require explicit state, provenance, tool preconditions, and authority boundaries. We ask: \emph{How can an onboard agent coordinate crew and robot sensing to reduce diagnostic uncertainty while keeping evidence traceable and the crew in control?} We contribute (1) an architecture linking anomaly events, crew interaction, robot inspection, hypothesis refinement, and crew-authorized action; (2) a bounded tool-use pattern separating information gathering from operational action; and (3) two scenarios showing the same evidence loop in a Mars-transit spacecraft and a lunar surface habitat. Here, \emph{triage} denotes decision support under unresolved uncertainty, not autonomous anomaly detection or emergency management.

\section{Evidence-Driven Agentic Architecture}

\subsection{Operational Scope and Explicit State}

The architecture continuously receives structured events from caution-and-warning systems, conventional fault detection, isolation, and recovery (FDIR), anomaly detectors, and crew reports. An off-nominal event initiates triage; independently designed automated safing may act in parallel when delay is unacceptable. The architecture then supports situation assessment and early response, when additional evidence can distinguish plausible causes and guide the next safe action. Here, \emph{agentic} denotes high-level coordination among finite, evidence-acquisition, context-retrieval, and communication capabilities exposed through typed interfaces and governed by explicit authority constraints, rather than unconstrained direct control of vehicle behavior.

The triage state manager owns a time-indexed record containing the initiating event, ranked hypotheses and uncertainty, evidence, operational context, robot and task status, applicable procedure context, and an append-only decision trace. Each evidence item records its source, modality, time, location, quality, provenance, and underlying artifact. Missing evidence, failed acquisition, and a measured-negative result remain distinct; dialogue history is not authoritative vehicle state.

\subsection{Fixed Roles and Typed Interfaces}

The architecture follows a hub-and-spoke workflow. A central \textbf{agentic orchestrator} receives structured anomaly notifications, manages dialogue through the crew-facing embodied agent, queries triage state, and routes typed requests and results among bounded services. It may sequence candidate evidence-acquisition actions and explanations, but it does not own authoritative state, assign diagnostic confidence, validate its own authority, execute critical actions, or issue low-level robot commands. \textbf{Event and monitoring services} emit structured anomaly events and immutable observations. The \textbf{triage state manager} owns the authoritative state, admits validated evidence, checks freshness and conflict, and enforces state transitions. The \textbf{crew-facing embodied agent} renders alerts and questions, confirms crew reports by readback, and explains assessment changes. Diagnostic, evidence, context/procedure, perception, and robot-tasking services expose bounded functions through typed interfaces.

A \textbf{mobile inspection robot} receives a typed task contract specifying the target, sensing objective, permitted modalities, completion and stop conditions, and timeout, then returns status, localized observations, and quality flags. An \textbf{authority and safety layer} validates schemas, permissions, operational constraints, and confirmation requirements for every request. During crewed operations, consequential actions require crew authorization; during uncrewed periods, only pre-authorized, reversible actions within verified constraints may execute automatically. Delayed ground may review the trace and provide advisory analysis when communications permit.

\subsection{Closed Evidence Loop and Authority Boundaries}

When an event arrives, the orchestrator asks the state manager to instantiate or retrieve the corresponding state and requests an initial diagnostic update. From the unresolved distinctions, it proposes a bounded evidence action: a crew question, data or context request, or robot inspection. The authority layer validates the request and obtains any required confirmation. The selected service executes it, the evidence service checks the result for schema, provenance, freshness, and quality, and the state manager admits it to the authoritative record. A diagnostic service revises the hypothesis ranking, and the crew-facing interface explains the supporting evidence, remaining uncertainty, and response options. The crew may accept, modify, decline, or defer; the trace remains available to delayed ground.

The architecture distinguishes \emph{epistemic actions}, which gather information, from \emph{operational actions}, which alter vehicle configuration or commit the crew to a consequential response. The orchestrator selects epistemic actions only from a bounded catalog with typed preconditions and outputs. Operational actions use a deterministic execution path and require crew authorization except for pre-approved safing. Capability access therefore does not confer operational authority, and the orchestrator cannot declare resolution without an appropriate validated measurement and operational approval.

\begin{figure*}[t]
\centering
\includegraphics[width=\textwidth]{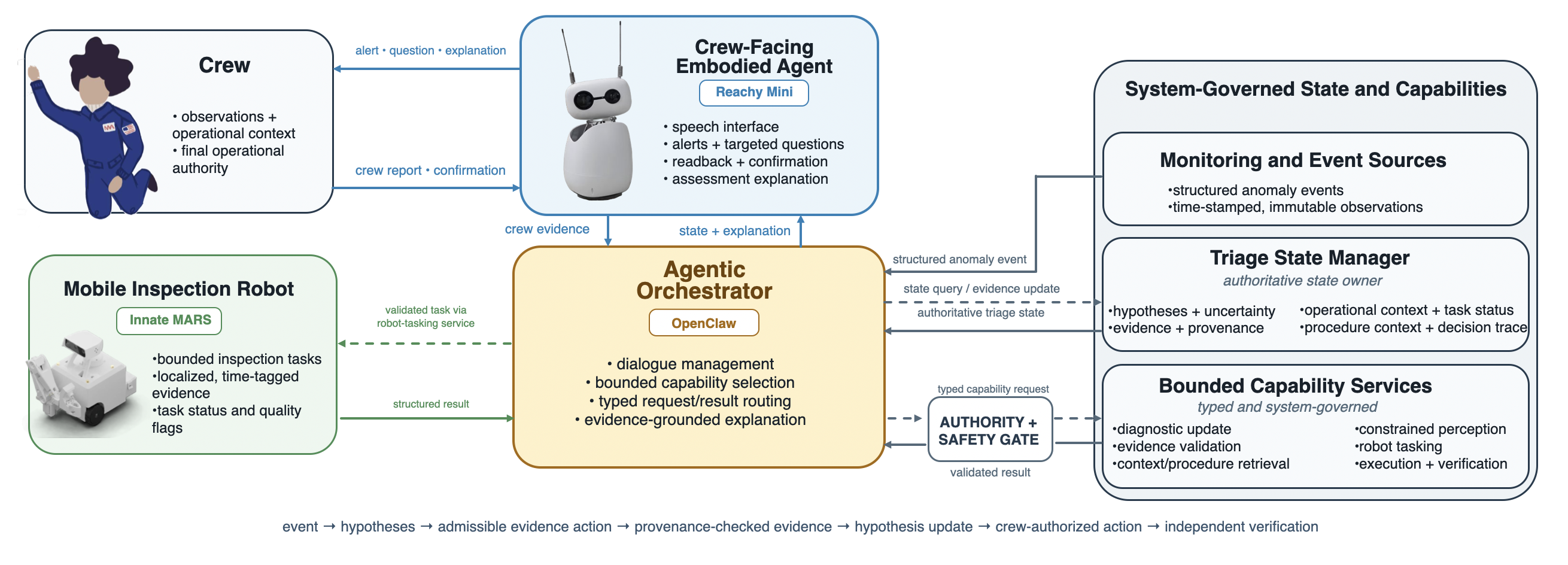}
\caption{Human--Agent--Robot architecture and proof-of-concept components. The Agentic Orchestrator coordinates the crew interface and mobile inspection robot with authoritative triage state and bounded services, while system-governed checks and crew authorization constrain operational actions.}
\label{fig:architecture}
\end{figure*}

\section{Illustrative Use Cases}

\subsection{Suspected Ammonia Leak During Mars Transit}

This scenario uses an actual ISS event to examine how a similar anomaly could be handled aboard a crewed vehicle in transit to Mars, where communication delay prevents real-time support from Mission Control. On January 14, 2015, the ISS annunciated \emph{TOXIC ATMOSPHERE Node 2 LTL IFHX NH3 Leak Detected}, indicating possible ammonia ingress through the Node 2 Low Temperature Loop Interface Heat Exchanger (IFHX). Software inhibited ventilation, and the crew isolated in the Russian Segment. Because expected confirming cues were absent, controllers traced the alarm to an erroneous Node 2-2 multiplexer/demultiplexer (MDM) indication. Power-cycling the unit cleared the anomaly, and atmosphere measurements detected no ammonia \cite{nasa2015,nasaoutpost2017}. We use the event only as an operational anchor, and the adaptation does not reproduce an official procedure.


\begin{enumerate}\setlength{\itemsep}{1pt}\setlength{\parskip}{0pt}\setlength{\parsep}{0pt}

\item \textbf{Alarm and safing.} A toxic-atmosphere indication reports possible ammonia ingress through an IFHX in a crew module. Pre-approved logic isolates the affected ventilation paths. Through the crew-facing interface, the orchestrator informs the crew: ``Possible ammonia ingress detected. Ventilation has been isolated. Move to the designated safe area and confirm when all crew are clear.''

\item \textbf{Crew evidence.} After the crew confirms its location and isolation status, the orchestrator asks whether anyone observed irritation, haze, droplets, unusual noise, or abnormal equipment behavior before leaving. The original responses and their structured interpretation are recorded as evidence without requiring crew re-entry.

\item \textbf{Preliminary assessment.} The diagnostic service ranks three hypotheses: genuine ammonia ingress through the IFHX, an erroneous sensing or data-handling indication, and another thermal-control anomaly. The absence of crew-observed cues weakens, but does not eliminate, the genuine-leak hypothesis. Because its consequence remains severe, isolation stays in effect.

\item \textbf{Robot inspection.} The diagnostic service requests a localized environmental evidence to further discriminate the hypothesis. After authority validation, the orchestrator dispatches a bounded task to an intravehicular free-flyer stationed in the isolated module. The task specifies the IFHX region, requested measurements, success and failure conditions, and timeout. An RGB camera, thermal imager, and payload-mounted NH$_3$/air-quality sensor collect time- and location-tagged evidence.

\item \textbf{Hypothesis update.} The robot reports no NH$_3$ detection and no visible or thermal abnormality, while separately retrieved diagnostics show an inconsistent state in the local MDM or data-handling unit. The diagnostic service consequently ranks an erroneous instrumentation indication as the leading hypothesis. The orchestrator explains which observations produced the change while emphasizing that a measured negative at the inspection location does not establish cabin safety.

\item \textbf{Crew-authorized verification.} The procedure service retrieves a validated MDM power-cycle sequence. After explicit crew authorization, a deterministic execution path performs the reset and monitoring confirms that the anomalous telemetry has cleared. The orchestrator reports that the evidence supports an instrumentation fault but does not declare the module safe; recovery requires an independent cabin-atmosphere measurement and crew confirmation.

\end{enumerate}

\subsection{Lunar Surface Power-Interface Anomaly}

NASA's Phase Three Moon Base concept envisions sustained human presence supported by habitats, power and communications infrastructure, logistics, and surface robotic assets \cite{nasamoonbasesystems2026,nasamoonbaseguide2026}. Because power is shared across habitation, mobility, science, and in situ resource utilization (ISRU), local interface faults are operationally important and may be difficult to distinguish using remote telemetry alone.

\begin{enumerate}
\setlength{\itemsep}{0pt}
\setlength{\topsep}{2pt}
\setlength{\partopsep}{0pt}
\setlength{\parskip}{0pt}
\setlength{\parsep}{0pt}

\item \textbf{Detection and context.} A surface power manager reports unstable voltage or current and abnormal interface temperature. The interface communicates the affected node and safing state, then obtains recent context such as connector mating, power switching, robotic work, cable deployment, or regolith disturbance.

\item \textbf{Initial assessment.} The diagnostic service ranks a source- or load-side fault, a connector or contact fault, and cable damage or mechanical displacement.

\item \textbf{Robot inspection.} After authority validation, an available surface robot receives a bounded task to collect close-range RGB and thermal imagery of the connector and adjacent cable, with validated local electrical measurements when available.

\item \textbf{Evidence update and response.} A partially seated, regolith-contaminated connector with localized heating and a nominal adjacent cable makes an interface fault the leading hypothesis. The state manager records the update, and the interface presents bounded options: isolate the node, reroute the load, execute approved robotic remediation, or schedule crew servicing. During uncrewed operation, only pre-authorized reversible actions may execute.

\item \textbf{Verification.} Power telemetry and robot reinspection determine whether transfer has stabilized and temperature is decreasing. Otherwise, the node remains isolated and the evidence loop continues.

\end{enumerate}

\section{Proof-of-Concept Implementation and Demonstration}

We implemented the Section~II architecture on an NVIDIA DGX Spark for the Section~III-A ammonia scenario. An OpenClaw-based orchestrator routes high-level workflow messages among Reachy Mini as the crew-facing interface, an authoritative triage state manager and bounded backend services, and Innate MARS through a ROS2 robot-tasking service \cite{reachymini,innatemars}. Local models provide dialogue and tool use (\texttt{Gemma-4-26B-A4B-it}), speech recognition and synthesis (\texttt{Parakeet-TDT-0.6B-v2}, \texttt{Kokoro-82M}), and visual interpretation (\texttt{Qwen3-VL-8B-Instruct}). Inter-component coordination uses structured state and typed messages; free-form language is exposed only at the crew interface.

\subsection{Implemented System}

The orchestrator manages dialogue, proposes capability calls, and generates evidence-grounded explanations, while deterministic services own state, compute hypothesis rankings, validate requests, retrieve procedures, and execute consequential actions. Reachy provides speech, alerts, readback, and explanation. MARS autonomously navigates to a mock IFHX/rack target and returns physical imagery together with a simulated NH$_3$ reading. The flight scenario assumes an intravehicular free-flyer; MARS serves as a morphology-independent surrogate for exercising the same task, status, and evidence interfaces, not for validating free-flight mobility. Source images remain primary evidence, while schema-constrained visual interpretations are stored as derived observations with model and artifact metadata.

\subsection{End-to-End Demonstration}

\begin{enumerate}\setlength{\itemsep}{1pt}\setlength{\parskip}{0pt}\setlength{\parsep}{0pt}

\item \textbf{Trigger and initial assessment.} Synthetic spacecraft telemetry generates an \texttt{AnomalyEvent} containing the affected subsystem, triggering measurements, severity, timestamp, and vehicle state. The orchestrator routes it to the diagnostic service, which ranks a genuine ammonia release, an erroneous instrumentation indication, and a non-toxic thermal-control transient. The resulting \texttt{HypothesisState} is stored by the triage state manager.

\item \textbf{Crew interaction.} Through Reachy Mini, the orchestrator alerts the crew, confirms isolation, and explains that the alarm remains diagnostically ambiguous. It then requests observations such as irritation, haze, droplets, or unusual equipment behavior. Local speech processing captures the response, and both the original utterance and structured \texttt{CrewObservation} are stored with provenance. The diagnostic service updates the ranking but retains the release hypothesis because the crew observations alone cannot clear the hazard.

\item \textbf{Evidence-driven inspection.} The diagnostic service identifies a localized environmental observation as the next discriminator, while action validation rejects direct crew entry. The orchestrator issues a bounded \texttt{InspectionRequest} containing the semantic IFHX/rack target, requested observations, diagnostic objective, and termination conditions. The robot-tasking service resolves the target to a predefined map pose, and MARS navigates autonomously while the orchestrator remains available to the crew.

\item \textbf{Structured evidence return.} At the target, MARS captures an image and obtains a simulated localized NH$_3$ measurement. The perception service uses the vision-language model only to produce a constrained visual observation; the raw image remains the primary evidence artifact. MARS returns task status, target, image reference, structured observation, NH$_3$ result, timestamp, and quality flags as an \texttt{InspectionResult}. The evidence service distinguishes this measured negative result from missing data or a failed acquisition.

\item \textbf{Refinement and crew-gated verification.} The diagnostic service integrates telemetry, crew reports, and robot evidence, shifting the leading hypothesis toward an instrumentation fault. Reachy explains the evidence responsible for the update while maintaining isolation. The procedure service then returns a predefined verification sequence. After explicit crew approval, a deterministic tool resets the mock instrumentation component; monitoring independently confirms that the telemetry has cleared, and a separate verification input reports no NH$_3$. The crew retains authority over the recovery transition.
\end{enumerate}

In the demonstrated run, the complete sequence executes end to end. The append-only trace records each state transition, tool request, authority check, robot result, explanation, and crew confirmation. This demonstrates integration of the closed evidence loop and its authority boundaries.

\section{Discussion and Conclusion}

Our contribution is an integrated, traceable, authority-bounded evidence loop for onboard anomaly triage. It connects an unresolved diagnostic question to targeted crew or robot sensing, hypothesis revision, and a crew-authorized response while keeping observations, inferences, and actions explicit. The two scenarios illustrate how a mobile sensing platform can extend onboard anomaly triage by gathering evidence from a suspected hazardous volume or inspecting a physical interface not directly observable through telemetry.

The hardware-in-the-loop prototype demonstrates end-to-end system integration, not improvement in diagnostic or human-performance outcomes; quantitative evaluation remains future work. Future work will focus on connecting validated diagnostic and procedure services, expanding multimodal sensing, and testing missing, delayed, and conflicting evidence. Habitat-analog studies, including NASA's Crew Health and Performance Exploration Analog (CHAPEA), could measure diagnostic accuracy, workload, trust, and decision quality under reduced ground support \cite{nasachapea}.

\end{document}